\documentclass{article}
\usepackage{ijcai24}

\usepackage{times}
\usepackage{soul}
\usepackage{url}
\usepackage[hidelinks]{hyperref}
\usepackage[utf8]{inputenc}
\usepackage[small]{caption}
\usepackage{graphicx}
\usepackage{amsmath}
\usepackage{amsthm}
\usepackage{booktabs}
\usepackage{algorithm}
\usepackage{algorithmic}
\usepackage[switch]{lineno}
\usepackage{multirow}
\usepackage{subcaption}
\usepackage{cleveref}
\usepackage{hyperref}
\usepackage{float}

\newcommand\algorithmicprocedure{\textbf{procedure}}
\newcommand{\algorithmicendprocedure}{\algorithmicend\ \algorithmicprocedure}
\makeatletter
\newcommand\PROCEDURE[3][default]{%
  \ALC@it
  \algorithmicprocedure\ \textsc{#2}(#3)%
  \ALC@com{#1}%
  \begin{ALC@prc}%
}
\newcommand\ENDPROCEDURE{%
  \end{ALC@prc}%
  \ifthenelse{\boolean{ALC@noend}}{}{%
    \ALC@it\algorithmicendprocedure
  }%
}
\newenvironment{ALC@prc}{\begin{ALC@g}}{\end{ALC@g}}
\makeatother

\title{Toward Lifelong Learning in Equilibrium Propagation: Sleep-like and Awake Rehearsal for Enhanced Stability
}

\author{
Yoshimasa Kubo
\and
Jean Erik Delanois \and
Maxim Bazhenov\\
\affiliations
Department of Medicine, University of California San Diego, La Jolla, USA\\
\emails
\{ykubo, jdelanoi, mbazhenov\}@ucsd.edu
}
\begin{document}

\maketitle

\begin{abstract}
Recurrent neural networks (RNNs) trained using Equilibrium Propagation (EP), a biologically plausible training algorithm, have demonstrated strong performance in various tasks such as image classification and reinforcement learning. However, these networks face a critical challenge in continuous learning: catastrophic forgetting, where previously acquired knowledge is overwritten when new tasks are learned. This limitation contrasts with the human brain's ability to retain and integrate both old and new knowledge, aided by processes like memory consolidation during sleep through the replay of learned information. To address this challenge in RNNs, here we propose a sleep-like replay consolidation (SRC) algorithm for EP-trained RNNs. We found that SRC significantly improves RNN's resilience to catastrophic forgetting in continuous learning scenarios. 
In class-incremental learning with SRC implemented after each new task training, the EP-trained multilayer RNN model (MRNN-EP) performed significantly better compared to feedforward networks incorporating several well-established regularization techniques. The MRNN-EP performed on par with MRNN trained using Backpropagation Through Time (BPTT) when both were equipped with SRC on MNIST data and surpassed BPTT-based models on the Fashion MNIST, Kuzushiji-MNIST, CIFAR10, and ImageNet datasets. Combining SRC with rehearsal, also known as "awake replay", further boosted the network’s ability to retain long-term knowledge while continuing to learn new tasks. Our study reveals the applicability of sleep-like replay techniques to RNNs and highlights the potential for integrating human-like learning behaviors into artificial neural networks (ANNs).
\end{abstract}

\section{Introduction}
Backpropagation (BP) has revolutionized the field of neural networks, enabling remarkable achievements in areas such as image classification \cite{Krizhevsky2012,Simonyan2014,Szegedy2015}, strategic game-play in Go \cite{Silver2016,Silver2017}, and mastery of Atari games \cite{Mnih2013,Mnih2015}. Despite its success, BP is not biologically plausible \cite{Lee2015,Whittington2019,Lillicrap2020,Shervani2023} and struggles with catastrophic forgetting, a key limitation in continual learning where previously acquired knowledge is overwritten when new tasks are introduced \cite{vandeVen2024}.

Equilibrium Propagation (EP) \cite{Scellier2017,Scellier2019,Ernoult2019,Laborieux2021,Laborieux2022} offers a biologically plausible alternative for training energy-based recurrent neural networks (RNNs), particularly convergent RNNs. EP operates through two distinct phases: the free phase and the weakly clamped phase. During the free phase, neural activations evolve without any teaching signals, allowing the network to autonomously infer target outputs. In the weakly clamped phase, subtle teaching signals adjust the activations, nudging the output towards desired targets. Weight updates are calculated based on differences in activations between these two phases.

Recent advancements by Luczak et al. \citeyear{Luczak2022a} demonstrate that steady-state activity in the free phase can often be approximated by early neural dynamics, reducing the need for complete equilibration and enhancing the biological plausibility of EP. Furthermore, enhanced EP approaches \cite{Laborieux2021} have shown that convolutional networks trained with EP can achieve comparable performance to those trained using BP, underscoring EP potential as a biologically inspired training method.

\begin{figure}[ht]
    \centering
    \includegraphics[width=0.50\textwidth]{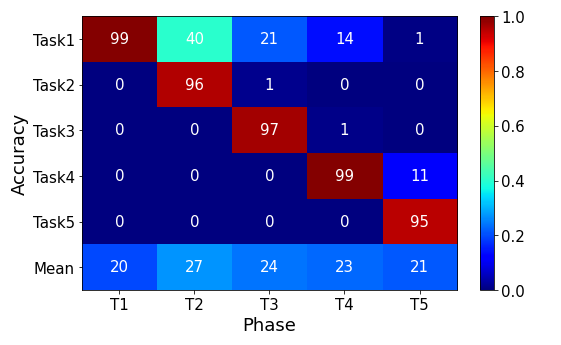}
    \caption{Performance of an RNN (one hidden layer with 1024 neurons) trained with EP on class-incremental MNIST. The y-axis indicates the tasks the RNN is tested on, while the x-axis represents the training phases. For example, "Task 1" on T1 corresponds to the RNN being trained on Task 1 (labels 0 and 1 of MNIST) during Training Phase 1, with accuracy measured across all tasks. Catastrophic forgetting occurs after training on new tasks, highlighting a limitation of EP despite its biological plausibility. For clarity, we use T\# to denote the sequential training phase number (e.g., T2 is the training phase for Task 2), and S\# to denote the sleep-like replay phase that follows that training phase (e.g., S2 follows T2).}
    \label{fig:rnn_wo_sleep}
\end{figure}

However, networks trained with EP remain vulnerable to catastrophic forgetting (Figure \ref{fig:rnn_wo_sleep}). Unlike the human brain, which can consolidate past knowledge while learning new information, EP-trained networks lack mechanisms akin to memory consolidation during sleep. 
While EP provides a biologically plausible framework, it requires additional mechanisms to emulate the robustness of biological learning systems.

A key hypothesis suggests that sleep in mammals supports generalization by replaying experiences from wakefulness \cite{Stickgold2013,Rasch2013,Lewis2011}. Feedforward neural networks trained with BP have been shown to mitigate catastrophic forgetting through sleep-like unsupervised replay \cite{Tadros2022}. Moreover, evidence suggests that the brain also engages in experience replay during awake states \cite{Foster2006,Diba2007,Carr2011,Findlay2020}. These insights inspire incorporating awake and sleep replay mechanisms into artificial neural networks.

Existing work on recurrent neural networks in continual learning \cite{Ehret2020,Cossu2021} has focused predominantly on sequential inputs rather than static datasets. In this study, we extend the paradigm of sleep-like replay to RNNs trained with EP, aiming to address catastrophic forgetting while preserving EP’s biological plausibility. This work advances the field by introducing biologically inspired solutions to key challenges in continual learning and demonstrates the potential of EP-trained networks to emulate human-like learning mechanisms.






\section{Methods}
In this section, we describe the methods used in our experiments. We focus on the implementation and training of recurrent neural networks using Equilibrium Propagation, the predictive learning rule, rehearsal techniques, and sleep replay consolidation. These methods are outlined to ensure reproducibility and to support comparisons in future research.

\subsubsection{Equilibrium Propagation (EP)}
Equilibrium Propagation (EP) is a biologically plausible learning algorithm that trains networks by minimizing an energy function through local, contrastive updates~\cite{Scellier2017,Ernoult2019}. This approach specifically leverages \textit{convergent recurrent neural networks}, whose internal dynamics are designed to settle into a stable, fixed-point state. The energy function $E$ defines the network’s internal configuration: neurons adjust their activations to reach states that locally minimize $E$. The state update dynamics follow the negative gradient of the energy function with respect to the neuron activations.

The convergent property is crucial for EP. It ensures that when a static input is provided, the network’s recurrent dynamics evolve over time until the neuron activations stop changing and reach a stable equilibrium. This convergence to an energy minimum underlies how EP generates learning signals using only local information.

In continuous time, the neuron states follow gradient flow:
\begin{equation}
    \frac{ds}{dt} = -\frac{\partial E}{\partial s}.
    \label{eq:continuous_dynamics}
\end{equation}
This means the neuron activations evolve continuously to reduce the total energy. The discretized version of this flow aligns with the update formulation described by Lin et al.~(2024) for energy-based recurrent dynamics, providing a clear link to modern formulations of equilibrium models.

EP training consists of two distinct phases. In the \textit{free phase}, the network settles naturally into a local energy minimum given an input, without any external target. In the \textit{weakly clamped phase}, a small nudging term gently pushes the output units towards the desired target, perturbing the equilibrium slightly. By contrasting the network states between these two conditions, synaptic weights are updated through purely local Hebbian-like rules.

\paragraph{Energy and Cost Relationship.}
To incorporate supervised learning, the total objective combines the internal energy $E$ and an external cost function $C$ that measures output error (e.g., squared difference from the target). During the weakly clamped phase, the dynamics minimize:
\begin{equation}
    F = E + \beta C,
    \label{eq:augmented_energy}
\end{equation}
where $\beta$ is a small nudging parameter. This coupling ensures that the output units are biased to reduce the cost, while the system still performs local energy minimization, producing a contrastive signal that approximates backpropagation but with local updates.

\paragraph{Energy Function.}
The network’s state dynamics are governed by minimizing an energy function $E(s, x, w)$, where $s$ denotes neuron activations, $x$ is the input, and $w$ are the synaptic weights. For example, a Hopfield-like recurrent network can be described by:
\begin{equation}
    E(s) = \frac{1}{2} \sum_{i} s_i^2 - \sum_{i<j} w_{ij} s_i s_j - \sum_{i} b_i s_i.
    \label{eq:energy_function}
\end{equation}
This structure encourages the network to settle into stable configurations that minimize $E$.

\paragraph{Free Phase.}
In the free phase, the network relaxes to a local minimum of the energy function given an input $x$. The continuous-time dynamics (Eq.~\ref{eq:continuous_dynamics}) are discretized using an Euler method. For a simple multilayer setting with one hidden layer:
\begin{equation}
    s_{o,t} = s_{o,t-1} + dt\, \Big(-\frac{\partial E}{\partial s_o}\Big), 
    \quad
    s_{j,t} = s_{j,t-1} + dt\, \Big(-\frac{\partial E}{\partial s_j}\Big).
\end{equation}
These expand to:
\begin{equation}
    s_{o,t} = s_{o,t-1} + dt\, \Big(-s_{o,t-1} + \sigma \big(\sum_j w_{jo} s_{j,t-1} + b_o \big)\Big),
    \label{eq:free_1}
\end{equation}
\begin{equation}
  \begin{split}
    s_{j,t} & = s_{j,t-1} + dt\, \Big(-s_{j,t-1} + \sigma \big(\sum_i w_{ij} x_i \\
            & + \gamma \sum_o w_{oj} s_{o,t-1} + b_j \big)\Big),
    \label{eq:free_2}
  \end{split}
\end{equation}
where $\sigma(\cdot)$ is an activation function, $\gamma$ is a feedback scaling parameter, and $dt$ is the integration step size.

\paragraph{Weakly Clamped Phase.}
In the weakly clamped phase, the output neurons receive a small nudging force that biases them towards the target $y$ by adding the cost gradient:
\begin{equation}
    \beta \frac{\partial C}{\partial s_o} = \beta (y - s_{o,t-1}),
    \label{eq:grad_c}
\end{equation}
assuming $C$ is mean squared error. The output dynamics become:
\begin{equation}
  \begin{split}
    s_{o,t} & = s_{o,t-1} + dt\, \Big(-s_{o,t-1} + \sigma \big(\sum_j w_{jo} s_{j,t-1} + b_o \big) \\
    & + \beta (y - s_{o,t-1})\Big).
    \label{eq:w_clamped_1}
  \end{split}
\end{equation}
The hidden states evolve as in the free phase (Eq.~\ref{eq:free_2}).

\paragraph{Weight Update Rule.}
Finally, the synaptic weights are updated by contrasting neuron correlations between the free and weakly clamped phases:
\begin{equation}
    \Delta w_{ij} = \frac{\alpha}{\beta} (\hat{s}_i \hat{s}_j - \check{s}_i \check{s}_j),
    \label{eq:conv_update}
\end{equation}
where $\hat{s}$ denotes activations in the weakly clamped state, $\check{s}$ those in the free state, and $\alpha$ is the learning rate.

This local contrastive update approximates the true gradient of the cost with respect to the weights, demonstrating how EP can perform supervised learning using only local, biologically plausible signals~\cite{Scellier2017}.

\subsubsection{Predictive Learning Rule}
Luczak et al. \cite{Luczak2022a} explored a biologically inspired variant of Hebbian learning for Equilibrium Propagation (EP) and similar local training algorithms. The update rule they introduced is called the \textit{predictive learning rule}. This formulation arises naturally if one assumes that each neuron strives to maximize its own metabolic energy balance by predicting its future activity and adjusting synaptic strengths accordingly.

The predictive learning rule is formally expressed as:
\begin{equation}
    \begin{split}
     \Delta w_{pre,post}  & = \frac{\alpha}{\beta} (\hat{s}_{pre} \hat{s}_{post} - \hat{s}_{pre} \check{s}_{post}) \\
     & = \frac{\alpha}{\beta} \hat{s}_{pre} ( \hat{s}_{post} -  \check{s}_{post}),
    \label{eq:pred_lr}        
    \end{split}
\end{equation}
where:
\begin{itemize}
  \item $w_{pre,post}$ denotes the synaptic weight connecting the \textit{presynaptic} neuron ($pre$) to the \textit{postsynaptic} neuron ($post$).
  \item $\hat{s}_{pre}$ is the activity of the presynaptic neuron during the weakly clamped phase.
  \item $\hat{s}_{post}$ is the actual activity of the postsynaptic neuron in the weakly clamped phase.
  \item $\check{s}_{post}$ is the actual activity of the postsynaptic neuron in the free phase.
\end{itemize}

In the original version, Luczak et al. used the \textit{predicted postsynaptic activity}, denoted by $\tilde{s}_{post}$, in place of the free-phase activity ($\check{s}_{post}$). Here, the “predicted postsynaptic activity” means the neuron’s internal estimate of what its activation will be in the near future, based on its internal state and local signals. This predictive signal represents the neuron’s capacity to anticipate its own next equilibrium value, which is thought to align with observations of predictive coding in biological neurons.

Their results showed that the correlation between the predicted ($\tilde{s}_{post}$) and actual ($\check{s}_{post}$) postsynaptic activity during the free phase was nearly perfect ($R = 1 \pm 0.0001$ SD), demonstrating that the explicit prediction provides negligible additional signal under these conditions.

Follow-up studies \cite{Luczak2022b,Kubo2022a,Kubo2022b} validated that a simplified version of the predictive rule—where the predicted signal is replaced directly with the actual free-phase activation—works well in practical tasks, such as image classification and reinforcement learning, without degrading performance or biological plausibility.

Therefore, in this study, which focuses on isolating the effect of sleep-like replay in continual learning, we adopt the simpler form without the explicit prediction term. This choice helps clarify the role of replay while maintaining the learning process’s theoretical foundation in biologically plausible predictive coding.

\subsubsection{Sleep Replay Consolidation (SRC)}
Tadros et al. \citeyear{Tadros2022} introduced Sleep Replay Consolidation (SRC), a sleep-like, unsupervised replay mechanism originally designed for feedforward neural networks (FFs) trained with backpropagation. In their approach, the trained artificial neural network (ANN) is converted to a spiking neural network (SNN) with the same architecture. During the sleep phase, the SNN’s activity is driven by randomly distributed Poisson spiking input. For each input vector, the probability of assigning a spike (value of 1) to a given element (e.g., input pixel) is sampled from a Poisson distribution whose mean rate is set by the mean intensity of that input element across the training data. For example, a pixel that is consistently bright across training samples will have a higher probability of spiking compared to a pixel with lower mean intensity. 

During the sleep phase, the learning rule used is a local Hebbian plasticity mechanism similar to spike-timing-dependent plasticity (STDP): synaptic weights are strengthened when presynaptic spikes precede postsynaptic spikes and weakened when postsynaptic spikes occur without preceding presynaptic input. This local STDP update replaces the standard EP-based predictive update rule during replay because the network operates in the spiking domain. After the sleep replay phase, the updated SNN weights are mapped back to the ANN used for task learning.

In our work, we extend SRC from feedforward to recurrent neural networks (RNNs). Specifically, we apply SRC to both the output and hidden recurrent layers to enable replay of temporal dependencies within the hidden states. To support this, we introduce feedback connections (e.g., transposed hidden-to-output weights) that govern the membrane potential dynamics in a manner consistent with the convergent EP updates used during wake learning. When the membrane potential crosses a threshold, the neuron spikes, and weights are adjusted using the local STDP rule. This mechanism enables spontaneous replay of past patterns within recurrent layers, helping preserve previously learned tasks and reduce catastrophic forgetting.

In our training procedure, these phases are interleaved: the network is first trained on tasks using the EP-based learning rule in the awake phase, and then SRC with the STDP rule is applied during a sleep phase after each new task. This combined approach mirrors the biological interleaving of awake learning and sleep-dependent consolidation observed in the brain \cite{Tadros2022}.

Prior work has shown that SRC-equipped models increase representational sparsity, reduce class overlap, and improve robustness to noise and adversarial perturbations \cite{Tadros2019,Delanois2023,Bazhenov2024}. In this study, our recurrent SRC implementation demonstrates how these benefits can extend to convergent RNNs trained with Equilibrium Propagation.

\begin{algorithm}[tb]
    \caption{SRC algorithm}
    \label{alg:algorithm}
    \begin{algorithmic}[1] 
    \PROCEDURE{Sleep}{$W, input, scales, threshold$}  
        \STATE \qquad Initialize $v$ (voltage) = 0
        \STATE \qquad S $\leftarrow$ 0s
        activity
        \FOR{$ t =1,2,\ldots, T$} 
            \STATE $\triangleright$ T $-$ Time step duration of sleep
            \STATE \qquad $S(-1)$ $\leftarrow$ Convert input to Poisson-distributed spiking
            \FOR{$ l =1,2,\ldots, L$} 
                \STATE $\triangleright$ L $-$ number of layers
                \IF{(l == 1)} 
                    \STATE  //Compute voltage at the output layer at first
                    \STATE $v_t(l)$ $\leftarrow$ $v_{t-1}(l)$ + scales[l] $*$ W[l]  $*$ $S_{t-1}[l+1]$
                \ELSE                
                    \STATE //Compute voltage with feedback
                    \STATE $v_t(l)$ $\leftarrow$ $v_{t-1}(l)$ + scales[l]  $*$ (W[l]  $*$  $S_{t-1}[l+1]$ + W[l-1] $*$ $S_{t-1}[l-1]$) 
                \ENDIF
                \STATE //Propagate spikes
                \STATE $S_t[l]$ $\leftarrow$ $v_t(l)$ $>$ threshold(l)
                \IF{(t != 1)} 
                    \STATE //STDP
                    \algsetup{linenosize=\tiny}
                    \scriptsize
                    \STATE $W[l]$ = $\left\{\begin{array}{lr}
                              W[l] + inc , & if S_t[l] = 1 \& S_{t-1}[l+1] = 1\\
                              W[l] - dec , & if S_t[l] = 1 \& S_{t-1}[l+1] = 0\\
                              \end{array}\right\}$  
                \ENDIF
           
            \STATE $v_t(l) $($v_t(l)$ $>$ threshold) $=$ 0
            \ENDFOR
        \ENDFOR

    \ENDPROCEDURE

    \end{algorithmic}
\end{algorithm}

\subsubsection{Rehearsal Method}
Rehearsal methods improve continual learning by reintroducing examples from previously learned tasks during training. Prior studies \cite{van2020,Rebuffi2017,Hayes2019,Kemker2017} have shown that rehearsal can substantially reduce forgetting and improve performance in sequential learning settings. In this study, we implemented a simple rehearsal strategy that uses a small, fixed subset (e.g., 2\%) of previously seen data when training on a new task, for both our model and baseline models \cite{Hsu2018}. We intentionally chose not to use more complex rehearsal frameworks such as iCaRL \cite{Rebuffi2017}, since our goal was to isolate and analyze the fundamental effect of explicit data replay.

We frame this replay as an \emph{awake} rehearsal process, inspired by neuroscientific findings that certain types of replay occur in the brain even during wakefulness \cite{Foster2006,Diba2007,Carr2011,Findlay2020}. Although there are more biologically sophisticated rehearsal approaches \cite{van2020}, we focus here on this minimal version to establish a clear comparison with the sleep-like SRC mechanism. This design allows us to evaluate the combined effects of explicit awake rehearsal and spontaneous sleep-like replay on mitigating catastrophic forgetting and enhancing task retention.

\subsubsection{Model and Dataset Specification}
For our experiments, we used five benchmark datasets: MNIST \cite{Lecun1998}, FMNIST, KMNIST, CIFAR10, and ImageNet. For illustration, the MNIST dataset comprises 60,000 28$\times$28 grayscale images of handwritten digits (10 classes).

Our main architecture is a recurrent neural network trained using Equilibrium Propagation (MRNN-EP). For MNIST, we use 782 input neurons, 1024 hidden neurons, and 10 output neurons (782–1024–10). Hidden layers use ReLU activations; the output uses a hard sigmoid as in \cite{Laborieux2021}.

To benchmark MRNN-EP, we include two conventional RNN baselines:
\begin{enumerate}
    \item \textbf{MRNN-BPTT}: A convergent RNN trained with Backpropagation Through Time (BPTT) following \cite{Ernoult2019}, using the same architecture for fair comparison.
    \item \textbf{RNN-BPTT}: A standard RNN with recurrent connections only in the hidden layer, trained with BPTT.
\end{enumerate}

To test continual learning performance, we compare different training strategies:
\begin{itemize}
    \item \textbf{Sequential training}: Tasks are learned one after another without replay or regularization.
    \item \textbf{Sleep Replay Consolidation (SRC)}: Our proposed biologically-inspired sleep-like replay mechanism, applied after each task.
    \item \textbf{Rehearsal}: A small percentage (2\%) of previous task data is stored and interleaved during new task training.
    \item \textbf{SRC + Rehearsal}: Combining SRC with rehearsal replay.
    \item \textbf{Parallel training}: An upper bound where all tasks are trained jointly (not sequentially), to show the best possible accuracy.
\end{itemize}

\begin{figure}[ht]
    \centering
    \includegraphics[width=0.50\textwidth]{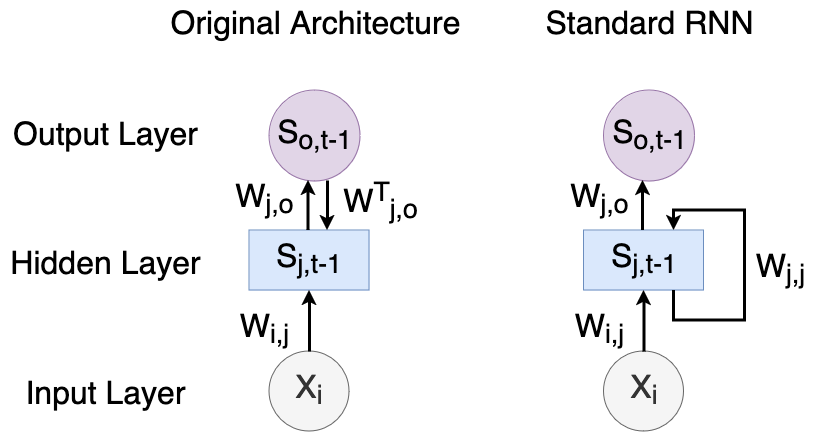}
    \caption{The architectures for our models and baselines. Left: The architecture for our convergent RNN model (MRNN-EP) and MRNN-BPTT incorporates a unique feedback weight structure. This feedback weight (transposed weights for the hidden to output layers, denoted as $W^T$ in the figure) connects the output layer to the hidden layer. This feedback mechanism, referred to as MRNN, distinguishes these networks from standard RNNs by enabling additional recurrent dynamics. Right: The architecture of the standard RNN trained with BPTT (RNN-BPTT) serves as a baseline model in our study. This architecture features recurrent connections solely within the hidden layer and lacks feedback weights that connect the output layer back to the hidden layer. As a result, the design is limited to sequential processing within the hidden layers, without the enhanced recurrent dynamics enabled by feedback mechanisms in MRNN models. This baseline provides a critical benchmark for evaluating the advantages of incorporating feedback dynamics in our experiments.}
    \label{fig:architech_rnn}
\end{figure}

Figure~\ref{fig:architech_rnn} illustrates the architectural differences between MRNN-EP and the standard RNN.


\textbf{Computational cost}: Using an NVIDIA GeForce RTX 3080 Ti, training MRNN-EP on five tasks takes approximately 4 minutes without SRC and 7 minutes with SRC — meaning that the full application of five sleep-like replay phases adds only ~3 minutes. This overhead is comparable to the cost of training a single task, indicating that SRC introduces only a modest increase in computational demand relative to the total training time, similar to previous work \cite{Tadros2022} .

\textbf{Phase Notation}: Throughout the paper — especially in the analysis sections — we use T\# to refer to the \#-th training phase, during which the model is trained on a new task, and S\# to denote the sleep-like replay phase that follows it. Importantly, the specific class labels assigned to each task may vary across experiments (e.g., to test stability under different task orders), so T2 does not always correspond to the same digit classes. This notation is used consistently in our confusion matrix visualizations, synaptic importance comparisons, and other analysis figures to indicate the position in the training sequence, rather than a fixed task identity.

Table~\ref{tbl:results_perform} summarizes the average test accuracy across tasks for all datasets, models, and training strategies.

\subsubsection{Fashion MNIST}
To evaluate performance on the more challenging Fashion MNIST dataset \cite{Xiao2017} (FMNIST), which consists of 60,000 28 $\times$ 28 grayscale images across 10 classes (e.g., T-shirt, Trouser, Pullover), we scaled the hidden layer size of our models to 2048 neurons. The overall architecture and activation functions remained identical to those used for MNIST. This adjustment accounts for the increased complexity of FMNIST compared to MNIST.



\subsubsection{Kuzushiji-MNIST}
To evaluate model performances on an additional dataset, we applied our model and the baseline models to the Kuzushiji-MNIST (KMNIST) dataset \cite{Lamb2018}. KMNIST consists of 60,000 grayscale images of handwritten Japanese Hiragana characters, each sized 28 $\times$ 28 pixels and categorized into 10 classes. The same architecture and activation functions used for the FMNIST dataset were applied to this dataset as well.

\subsubsection{CIFAR10}
All previously used datasets consisted of grayscale images. To assess performance on natural images, we tested our model and the baseline models on the CIFAR-10 dataset \cite{Krizhevsky2009}, which comprises 60,000 color images (32 $\times$ 32 pixels) across 10 object categories. Due to the higher dimensionality and complexity of this dataset, we extracted features using a convolutional network composed of VGG-style blocks \cite{Simonyan2014}, following the methodology of a previous study \cite{Tadros2022}.

The feature extractor architecture includes the following components:
\begin{itemize}
\item Block 1: Two convolutional layers with 64 kernels (3 $\times$ 3), each followed by batch normalization \cite{Ioffe2015} and ReLU activation.
\item Block 2: Two convolutional layers with 128 kernels (3 $\times$ 3), each followed by batch normalization and ReLU activation.
\item Block 3: Two convolutional layers with 256 kernels (3 $\times$ 3), each followed by batch normalization and ReLU activation.
\item Block 4: One convolutional layer with 512 kernels (3 $\times$ 3), followed by batch normalization and ReLU activation.
\end{itemize}

The feature vectors extracted from the output of this network were used as inputs to the learning models, each configured with a hidden layer containing 1024 neurons. The pretrained model in this experiment is from GitHub (VGG-13 with batch normalization at \url{https://github.com/chenyaofo/pytorch-cifar-models/tree/master?tab=readme-ov-file}).

\subsubsection{ImageNet}
To further assess model performance on a more complex natural image classification task, we applied the models to the ImageNet dataset \cite{Deng2009}. ImageNet contains 540,000 high-resolution color images (256 $\times$ 256 pixels) across 1,000 categories. For our experiment, we selected the first 10 classes (tench, goldfish, white shark, tiger shark, hammerhead, electric ray, stingray, cock, hen, and ostrich). Feature vectors were extracted from the top of the feature extraction layers (removing the classification layers) of a ResNet-152 model \cite{He2016} pretrained on ImageNet. These extracted features were then used as inputs for our learning models, which included a hidden layer with 1024 neurons. This pretrained model on ImageNet is available in PyTorch at (\url{https://pytorch.org/vision/main/models.html}).

All baseline models (MRNN-BPTT and RNN-BPTT) were trained using standard stochastic gradient descent (SGD) without advanced optimization techniques. For MNIST, FMNIST, and KMNIST, a fixed learning rate of 0.01 was used. For CIFAR-10 and ImageNet, learning rates of 0.023 and 0.02 were used for MRNN-BPTT and RNN-BPTT, respectively. The hyperparameters for our proposed model (MRNN-EP) are detailed in Table \ref{tbl:params_ep}. All models were trained using a batch size of 256. Training was performed for 3 epochs per task on MNIST, FMNIST, and KMNIST, and for 5 epochs per task on CIFAR-10 and ImageNet.

To assess upper-bound performance, we also trained all models on all tasks simultaneously ("parallel training" in Table \ref{tbl:results_perform}) for comparison with incremental training performance.

\subsubsection{Incremental Task Evaluation}
To test the efficacy of SRC in mitigating catastrophic forgetting, we created five incremental tasks from each dataset. Each task included data for two distinct labels (e.g., task 1 contained labels 0 and 1, task 2 contained labels 2 and 3, etc.) for all the datasets. To ensure consistency in evaluation, 10\% of each dataset was reserved for testing.

To assess the robustness of the models in continuous learning scenarios, we trained them six times on the five different tasks orders. The models' performance was then computed as an average across these training circumstances. This methodology provides valuable insights into the adaptability and generalization capabilities of the models, ensuring that the evaluation captures their performance consistency and ability to handle diverse task sequences effectively. Such an approach is crucial for understanding how well the models retain and integrate knowledge in continuous learning settings.

\subsubsection{SRC Hyperparameter Optimization}
The hyperparameters for SRC were optimized using a population-based Genetic Algorithm (GA) implemented in Python via the geneticalgorithm package \cite{solgi2020}. The GA was configured with a population size of 100, uniform crossover with a probability of 0.75, and a mutation probability of 0.1 per parameter. Elitism was applied by carrying over the top 1\% of individuals unchanged to the next generation, while 20\% of the population was selected as parents for reproduction using tournament selection.

To ensure robust and unbiased tuning, the GA used a held-out validation split comprising 10\% of the training data to evaluate the fitness of each candidate solution. The fitness function was defined as the average classification accuracy across all tasks on this validation set. The GA was run until convergence, with no fixed maximum number of generations.



\begin{table}[ht] \centering
\caption{The hyper-parameters for the MRNN-EP are summarized in this table. Here, $\alpha 1$ refers to the learning rate for updating the weights between the input and hidden layers, and $\alpha 2$ is the learning rate for updating the weights between the hidden and output layers. The 'Free Phase' and 'Clamped Phase' columns specify the number of time steps used during the free and weakly clamped phases, respectively. $\gamma$ is the feedback parameter, $\beta$ is a nudging parameter for the weakly clamped phase, and $dt$ is the Euler methods’ time-step. Batch and hidden sizes for baselines are the same as in this table.}
\resizebox{\columnwidth}{!}{%
\begin{tabular}{|l|c|c|c|c|c|}
\hline
Parameters    & MNIST & FMNIST & KMNIST & CIFAR10 & ImageNet \\ \hline
$\alpha 1$      & 0.03  & 0.03 & 0.03  & 0.08 & 0.08 \\
$\alpha 2$      & 0.001 & 0.001 & 0.001 & 0.001 & 0.001 \\
$\beta $      & 1.0   & 1.0    & 1.0 & 1.0 & 1.0 \\
$dt$          & 0.2   & 0.2  & 0.2 & 0.2  & 0.2 \\
$\gamma$      & 1.0   & 1.0  & 1.0  & 1.0 & 1.0 \\
Free phase    & 100   & 125  & 125 & 125 & 125  \\
Clamped phase & 10    & 15   & 15  & 15  & 15 \\
Hidden size   & 1024  & 2048 & 2048 & 1024 & 1024 \\
Batch size    & 256   & 256  & 256 & 256 & 256  \\
Epoch per task & 3   & 3  & 3 & 5 & 5 \\ \hline
\end{tabular}
}
\label{tbl:params_ep}
\end{table}

\begin{table*}[h!t]
\centering
\caption{
Average test accuracies ($\pm$ standard deviation) for all datasets under different training strategies: sequential training, Sleep Replay Consolidation (SRC), rehearsal with 2\% memory, SRC combined with rehearsal, and parallel training (joint learning of all tasks). Results are averaged across multiple task orders for robustness.
}
\begin{tabular}{|c|c|c|c|c|}
\hline
\textbf{Dataset} & \textbf{Method} & \textbf{MRNN-EP} & \textbf{MRNN-BPTT} & \textbf{RNN-BPTT} \\
\hline
\textbf{MNIST} & Sequential training & 24.76 $\pm$ 3.03\% & 20.88 $\pm$ 1.87\% & 20.43 $\pm$ 1.07\% \\
 & SRC & 64.68 $\pm$ 4.35\% & 65.41 $\pm$ 4.98\% & 49.88 $\pm$ 3.28\% \\
 & Rehearsal (2\% data) & 31.15 $\pm$ 1.99\% & 29.57 $\pm$ 0.65\% & 26.19 $\pm$ 2.01\% \\
 & SRC + Rehearsal (2\% data) & 67.77 $\pm$ 3.98\% & 66.92 $\pm$ 3.91\% & 55.29 $\pm$ 3.26\% \\
 & Parallel training & 96.59 $\pm$ 0.0008\% & 96.36 $\pm$ 0.0007\% & 97.33 $\pm$ 0.001\% \\
\hline
\textbf{FMNIST} & Sequential training & 21.17 $\pm$ 3.31\% & 21.91 $\pm$ 3.46\% & 22.50 $\pm$ 3.40\% \\
 & SRC & 45.73 $\pm$ 1.98\% & 37.92 $\pm$ 5.77\% & 33.21 $\pm$ 7.74\% \\
 & Rehearsal (2\% data) & 30.70 $\pm$ 4.09\% & 31.77 $\pm$ 4.10\% & 34.40 $\pm$ 5.42\% \\
 & SRC + Rehearsal (2\% data) & 50.90 $\pm$ 2.10\% & 43.90 $\pm$ 6.01\% & 39.62 $\pm$ 6.93\% \\
 & Parallel training & 87.53 $\pm$ 0.006\% & 88.16 $\pm$ 0.007\% & 87.83 $\pm$ 0.004\% \\
\hline
\textbf{KMNIST} & Sequential training & 19.94 $\pm$ 2.25\% & 18.95 $\pm$ 0.20\% & 19.10 $\pm$ 0.16\% \\
 & SRC & 48.39 $\pm$ 2.90\% & 43.13 $\pm$ 9.99\% & 39.88 $\pm$ 3.52\% \\
 & Rehearsal (2\% data) & 22.57 $\pm$ 2.91\% & 22.41 $\pm$ 3.06\% & 21.07 $\pm$ 1.50\% \\
 & SRC + Rehearsal (2\% data) & 50.27 $\pm$ 3.12\% & 46.89 $\pm$ 9.61\% & 41.98 $\pm$ 5.45\% \\
 & Parallel training & 94.97 $\pm$ 0.004\% & 94.88 $\pm$ 0.001\% & 95.71 $\pm$ 0.001\% \\
\hline
\textbf{CIFAR10} & Sequential training & 33.56 $\pm$ 3.19\% & 20.52 $\pm$ 0.83\% & 19.67 $\pm$ 0.52\% \\
 & SRC & 81.30 $\pm$ 1.98\% & 41.13 $\pm$ 2.93\% & 31.69 $\pm$ 3.02\% \\
 & Rehearsal (2\% data) & 43.53 $\pm$ 2.07\% & 31.07 $\pm$ 2.08\% & 28.39 $\pm$ 2.21\% \\
 & SRC + Rehearsal (2\% data) & 83.49 $\pm$ 1.39\% & 50.35 $\pm$ 1.33\% & 51.11 $\pm$ 2.11\% \\
 & Parallel training & 91.87 $\pm$ 0.0003\% & 91.93 $\pm$ 0.0009\% & 91.78 $\pm$ 0.003\% \\
\hline
\textbf{ImageNet} & Sequential training & 58.15 $\pm$ 3.78\% & 49.09 $\pm$ 4.57\% & 52.43 $\pm$ 3.60\% \\
 & SRC & 77.71 $\pm$ 2.91\% & 70.46 $\pm$ 2.43\% & 64.79 $\pm$ 3.37\% \\
 & Rehearsal (2\% data) & 67.07 $\pm$ 3.27\% & 60.78 $\pm$ 3.31\% & 65.70 $\pm$ 4.02\% \\
 & SRC + Rehearsal (2\% data) & 81.35 $\pm$ 2.10\% & 75.31 $\pm$ 1.55\% & 70.48 $\pm$ 3.11\% \\
 & Parallel training & 89.67 $\pm$ 0.003\% & 88.63 $\pm$ 0.006\% & 90.10 $\pm$ 0.007\% \\
\hline
\end{tabular}
\label{tbl:results_perform}
\end{table*}


\section{Results}
\begin{figure*}[ht]
  \begin{subfigure}{0.33\textwidth}
    \includegraphics[width=\linewidth, height=5cm]{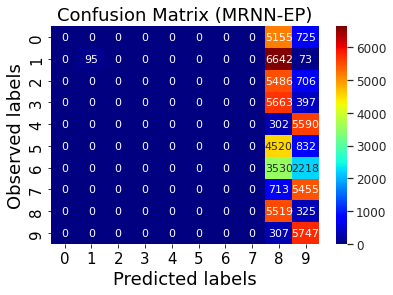}
    \caption{MRNN-EP before SRC}
    \label{fig:cm_ep_wo_src}
  \end{subfigure}%
  \hfill
  \begin{subfigure}{0.33\textwidth}
    \includegraphics[width=\linewidth, height=5cm]{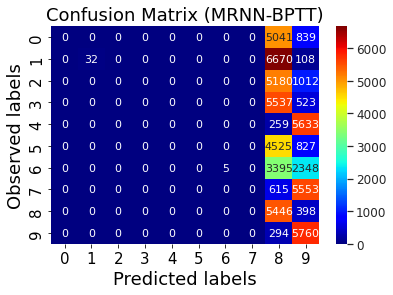}
    \caption{MRNN-BPTT before SRC}
    \label{fig:cm_bptt_wo_src}
  \end{subfigure}%
  \hfill
  \begin{subfigure}{0.33\textwidth}
    \includegraphics[width=\linewidth, height=5cm]{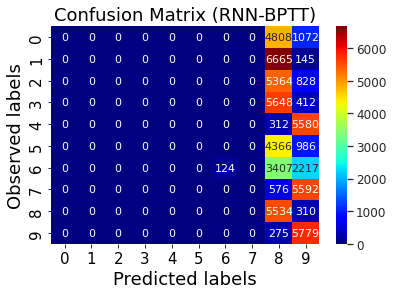}
    \caption{RNN-BPTT before SRC}
    \label{fig:cm_std_wo_src}
  \end{subfigure} \\

  \begin{subfigure}{0.33\textwidth}
    \includegraphics[width=\linewidth, height=5cm]{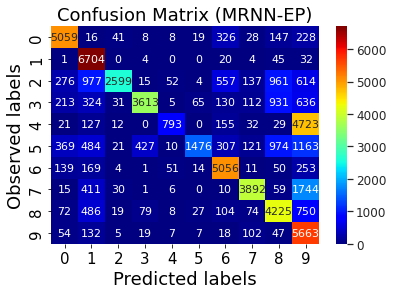}
    \caption{MRNN-EP after SRC}
    \label{fig:cm_ep_w_src}
  \end{subfigure}%
  \hfill
  \begin{subfigure}{0.33\textwidth}
    \includegraphics[width=\linewidth, height=5cm]{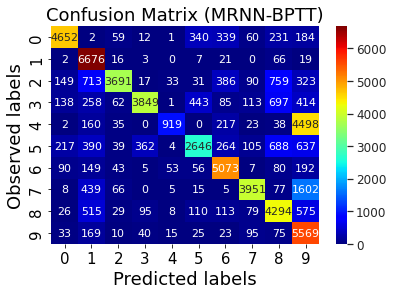}
    \caption{MRNN-BPTT after SRC}
    \label{fig:cm_bptt_w_src}
  \end{subfigure}%
  \hfill
  \begin{subfigure}{0.33\textwidth}
    \includegraphics[width=\linewidth, height=5cm]{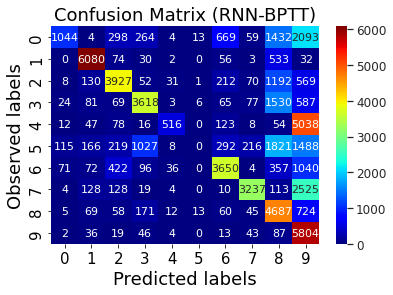}
    \caption{RNN-BPTT after SRC}
    \label{fig:cm_std_w_src}
  \end{subfigure} \\

    \caption{Confusion matrices for the MNIST dataset. \textbf{Top:} Model predictions immediately after sequential training on all tasks but \emph{before} applying the final sleep-like replay phase (i.e., after T5). \textbf{Bottom:} Predictions immediately \emph{after} the final replay phase (S5). Each matrix shows average results over 6 independent trials. Note that task label assignments may vary across trials to test model stability. These matrices illustrate how SRC helps recover correct class assignments degraded by catastrophic forgetting during sequential learning.}
  \label{fig:cm_mnist}
\end{figure*}

\begin{figure*}[ht]
  \begin{subfigure}{0.33\textwidth}
    \includegraphics[width=\linewidth, height=5cm]{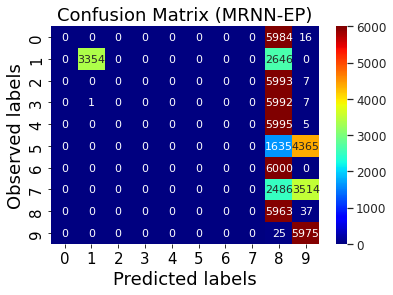}
    \caption{MRNN-EP before SRC}
    \label{fig:cm_ep_fm_wo_src}
  \end{subfigure}%
  \hfill
  \begin{subfigure}{0.33\textwidth}
    \includegraphics[width=\linewidth, height=5cm]{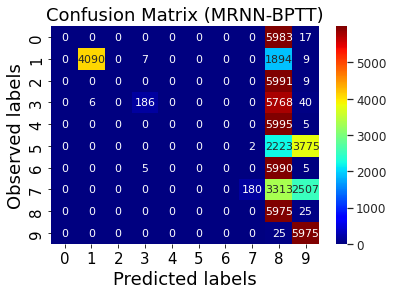}
    \caption{MRNN-BPTT before SRC}
    \label{fig:cm_bptt_fm_wo_src}
  \end{subfigure}%
  \hfill
  \begin{subfigure}{0.33\textwidth}
    \includegraphics[width=\linewidth, height=5cm]{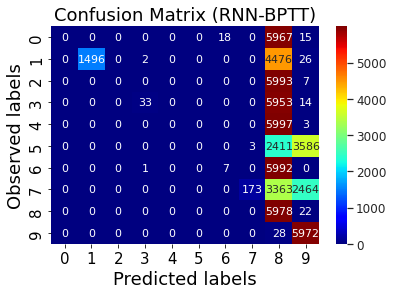}
    \caption{RNN-BPTT before SRC}
    \label{fig:cm_std_fm_wo_src}
  \end{subfigure} \\

  \begin{subfigure}{0.33\textwidth}
    \includegraphics[width=\linewidth, height=5cm]
    {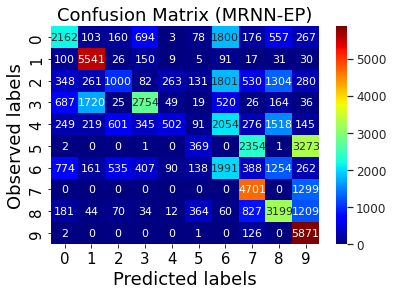}
    \caption{MRNN-EP after SRC}
    \label{fig:cm_ep_fm_w_src}
  \end{subfigure}%
  \hfill
  \begin{subfigure}{0.33\textwidth}
    \includegraphics[width=\linewidth, height=5cm]{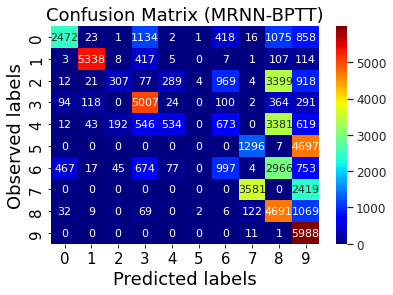}
    \caption{MRNN-BPTT after SRC}
    \label{fig:cm_bptt_fm_w_src}
  \end{subfigure}%
  \hfill
  \begin{subfigure}{0.33\textwidth}
    \includegraphics[width=\linewidth, height=5cm]{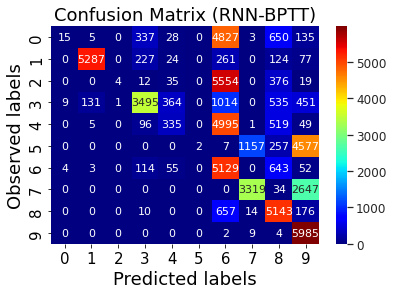}
    \caption{RNN-BPTT after SRC}
    \label{fig:cm_std_fm_w_src}
  \end{subfigure} \\

    \caption{Confusion matrices for the FMNIST dataset. \textbf{Top:} Model predictions immediately after sequential training on all tasks but \emph{before} applying the final sleep-like replay phase (i.e., after T5). \textbf{Bottom:} Predictions immediately \emph{after} the final replay phase (S5). Each matrix shows average results over 6 independent trials. As with MNIST, task order was varied across trials to evaluate model stability. The comparison demonstrates how SRC recovers class distinctions that were degraded by catastrophic forgetting during sequential learning.}

  \label{fig:cm_fmnist}
\end{figure*}

\subsubsection{Performances on MNIST}
Our key findings are summarized in Table \ref{tbl:results_perform}.
It compares the performance of our model (MRNN-EP) to two baselines — a MRNN-BPTT and a standard RNN-BPTT (SRC hyperparameters were tuned independently for each model) on the incremental MNIST dataset. In all cases, SRC was applied after each new task training.
Additionally, the ideal accuracies of each model are provided as "Parallel training" in the table. All results are obtained by averaging across multiple different task orders.

The results clearly demonstrate that integrating SRC significantly mitigated catastrophic forgetting
across all models. The performances for all the models improved by at least $ \sim 30\% $ when compared to RNNs trained without SRC. Notably, MRNN-EP with SRC achieved performance closely aligned with that of the MRNN-BPTT, reaching 64.68\% compared to 65.41\% for the MRNN-BPTT.

Next, we tested effect of combining SRC with rehearsal. As expected, rehearsal improved model performance, however, the gain was limited (about a 6\% improvement over baseline) when only 2\% of previously learned data was used (see below for detailed analysis of the effect of old dataset size). Combining the rehearsal method with SRC led to complimentary effect. Thus, the MRNN-EP with SRC using 2\% of old data achieved 67.77\% accuracy, closely rivaling the 66.92\% accuracy of the MRNN-BPTT. These findings underscore SRC's significant role when paired with rehearsal, driving substantial improvements in continual learning capabilities and narrowing the performance gap between EP-based models and those trained with traditional methods.

\begin{table*}[ht]
\centering
\caption{Average test accuracies ($\pm$ standard deviation) of MRNN-EP and FF-BP models on MNIST, FMNIST, and CIFAR10. SEQ denotes sequential training without replay. OWM results for FF-BP are from \protect\cite{Tadros2022}.}
\resizebox{0.85\textwidth}{!}{%
\begin{tabular}{|c|c|c|c|c|c|}
\hline
\multirow{2}{*}{Dataset} & MRNN-EP & MRNN-EP & FF-BP & FF-BP & FF-BP \\
                         & SEQ     & SRC     & SEQ   & SRC   & OWM   \\
\hline
MNIST                    & 24.76 $\pm$ 3.03\% & 64.68 $\pm$ 4.35\% & 19.49 $\pm$ 0.002\% & 48.47 $\pm$ 5.03\% & 77.04 $\pm$ 2.91\% \\
FMNIST                   & 21.17 $\pm$ 3.31\% & 45.73 $\pm$ 1.98\%  & 19.67 $\pm$ 0.003\% & 41.68 $\pm$ 5.04\% & 58.35 $\pm$ 2.05\% \\
CIFAR10                  & 33.56 $\pm$ 3.19\% & 81.30 $\pm$ 1.98\%  & 19.01 $\pm$ 0.002\% & 44.55 $\pm$ 1.45\% & 34.23 $\pm$ 1.87\% \\
\hline
\end{tabular}
}
\label{tbl:comp_ep_bp}
\end{table*}


\subsubsection{Performances on Fashion MNIST}
Having demonstrated the competitive performance of our models on the MNIST dataset, we extended our evaluation to the more challenging FMNIST dataset (Table \ref{tbl:results_perform}).

The results demonstrate that applying SRC led to significant performance improvements for all the models. Most notably, the MRNN-EP with SRC achieved an accuracy of 45.73\%, outperforming both the MRNN-BPTT (37.92\%) and the RNN-BPTT (33.21\%). These results underscore the effectiveness of SRC combined with EP in improving the model's ability to retain and generalize learned representations in more complex datasets like FMNIST.

As with the MNIST dataset, the application of the combination of SRC and the 'low-data' rehearsal method further boosted performance, leading to complementary model gains. When only 2\% of previously learned data was used, all the models implementing SRC+rehearsal showed at least a 5\% improvement over SRC alone. Among them, our MRNN-EP with SRC, coupled with the rehearsal method, achieved the highest performance of 50.90\%, highlighting its capacity to excel in continuous learning scenarios.

\subsubsection{Performances on Kuzushiji-MNIST}
The performance results for another grayscale task, KMNIST, are presented in Table \ref{tbl:results_perform}.

The results show that, once again, applying SRC improves the performance of all models. Specifically, the improvements for MRNN-EP, MRNN-BPTT, and RNN-BPTT are approximately 20\%, 15\%, and 20\%, respectively. Among these, MRNN-EP achieved the highest accuracy on this dataset as well.

However, for models incorporating both SRC and the rehearsal method, the performance gains were relatively modest—around 2–3\%—and similar to the improvements observed with rehearsal alone.

\subsubsection{Performance on CIFAR-10}
The performance of all models on the natural image classification task using the CIFAR-10 dataset is summarized in Table \ref{tbl:results_perform}. In this experiment, the input features were extracted using VGG-style blocks pretrained on CIFAR-10.

Under sequential training, MRNN-EP achieved an accuracy of 33.56\%, outperforming MRNN-BPTT (20.52\%) and RNN-BPTT (19.67\%). When SRC was applied, our model exhibited a substantial improvement of approximately 48\%, in contrast to 21\% for MRNN-BPTT and only 9\% for RNN-BPTT. Notably, our model with SRC reached 81.30\% accuracy, reducing the performance gap to only 10\% compared to parallel training (91.87\%). However, the addition of rehearsal to SRC (SRC+rehearsal) yielded only a modest improvement—approximately 2\%—indicating that SRC alone accounts for most of the performance gain.

\subsubsection{Performance on ImageNet}
We further evaluated the models on the ImageNet dataset, using extracted features from a pretrained ResNet-152 as input. The final row in Table \ref{tbl:results_perform} presents the results for this experiment.

All models performed relatively well under sequential training, achieving accuracies above 49\%. Specifically, MRNN-EP reached 58.15\%, while MRNN-BPTT and RNN-BPTT obtained 49.09\% and 52.43\%, respectively. One possible explanation for this relatively strong performance across models is that the extracted features from the pretrained ResNet-152 were highly separable, making the classification task easier.

The addition of SRC led to significant performance improvements for all models. MRNN-EP again achieved the highest accuracy (77.71\%), followed by MRNN-BPTT (70.46\%) and RNN-BPTT (64.79\%). Notably, applying SRC combined with rehearsal to our model further closed the gap with parallel training performance. The accuracy reached 81.35\%, leaving only an 8\% difference compared to the parallel training result (89.67\%), highlighting the effectiveness of our approach in mitigating forgetting even on complex, high-dimensional tasks.

\subsubsection{Comparison with Feedforward Neural Networks}
To evaluate the effectiveness of SRC in our recurrent setting, we compare MRNN-EP to a standard feedforward neural network trained with backpropagation (FF-BP). The results for FF-BP are adapted from \cite{Tadros2022}, including the baseline sequential training (SEQ), SRC, and Orthogonal Weight Modification (OWM) methods.

As shown in Table~\ref{tbl:comp_ep_bp}, MRNN-EP with SRC consistently outperforms FF-BP with SRC on all datasets, demonstrating that our recurrent extension of SRC yields stronger retention of past tasks. Compared to FF-BP with OWM—a strong regularization approach—MRNN-EP with SRC achieves comparable or better performance on MNIST and FMNIST, and notably surpasses it on the more challenging CIFAR-10 benchmark. This highlights that integrating sleep-like replay into a recurrent architecture can be as effective as well-established continual learning techniques for feedforward models.

To keep comparisons fair and relevant, we omit results for methods like EWC and SI, which are known to underperform in class-incremental scenarios without task labels.

In summary, these findings highlight that the EP-based RNN model, when enhanced with SRC—and further improved through rehearsal if old data are available—consistently outperforms BPTT-trained RNNs, particularly on more complex datasets. Moreover, MRNN-EP with SRC demonstrates strong competitiveness against feedforward neural networks equipped with state-of-the-art regularization techniques.
Crucially, it preserves the biological plausibility intrinsic to the EP framework, presenting a robust and effective solution for continuous learning tasks.

\subsubsection{Confusion Matrix Analysis}
To further compare the performance and error patterns of our models against baseline RNNs, we analyzed confusion matrices on the MNIST and FMNIST datasets. As discussed above, our MRNN-EP model achieves test accuracies comparable to the BPTT-trained RNN on MNIST, while on FMNIST the MRNN-EP with SRC demonstrates notably higher performance than both baselines.

To understand how SRC affects task-specific predictions, we examined the confusion matrices for MRNN-EP, MRNN-BPTT, and RNN-BPTT before and after the final SRC phase. Specifically, we visualized the confusion matrices immediately after sequential training on all tasks (T1 → S1 → T2 → S2 → T3 → S3 → T4 → S4 → T5), but prior to the final SRC replay phase (S5), and then again after S5 was completed. This comparison illustrates how SRC helps recover earlier task performance while preserving knowledge of newly learned tasks.

This comparison highlights how SRC recovers class-level accuracy that is otherwise lost due to catastrophic forgetting. Each matrix shown represents an average over six independent runs to ensure robustness of the results.

Before applying SRC, the confusion matrices revealed significant forgetting across all models, with predictions largely biased toward the most recent task (top of \Cref{fig:cm_mnist,fig:cm_fmnist}) on both datasets. This behavior highlights the susceptibility of these models to catastrophic forgetting. However, after applying SRC, the models successfully retained and recalled knowledge from previous tasks, clearly demonstrating SRC's effectiveness in mitigating forgetting (bottom of \Cref{fig:cm_mnist,fig:cm_fmnist}).

The confusion matrices provide detailed insights into the models' prediction patterns, revealing their strengths and weaknesses. Key observations include: 
\begin{enumerate} 
\item The confusion matrix for the MRNN-EP closely resembles that of the MRNN-BPTT, indicating similar error patterns across classes on the MNIST dataset. However, on the FMNIST dataset, MRNN-EP outperformed MRNN-BPTT, particularly for labels 5, 6, and 7.
\item The confusion matrix for the RNN-BPTT differs notably. This divergence can be attributed to the recurrent connections in the RNN’s hidden layer, which likely influence the formation of class representations and prediction patterns. 
\end{enumerate}

These findings reinforce the competitive performance of the MRNN-EP in continual learning scenarios, particularly on complex datasets like FMNIST, and highlight the influence of architectural differences on error patterns and prediction dynamics.

\subsubsection{Old data for rehearsal method with SRC}
Above, we report the effect of combining SRC with rehearsal using a limited amount (2\%)  of old data, and we  
consistently observed performance improvements across datasets. In this subsection, we systematically investigate the impact of varying the fraction of old data included in the rehearsal method on the model's performance on the MNIST dataset. 

\begin{figure}[ht]
    \centering
    \includegraphics[width=0.40\textwidth]{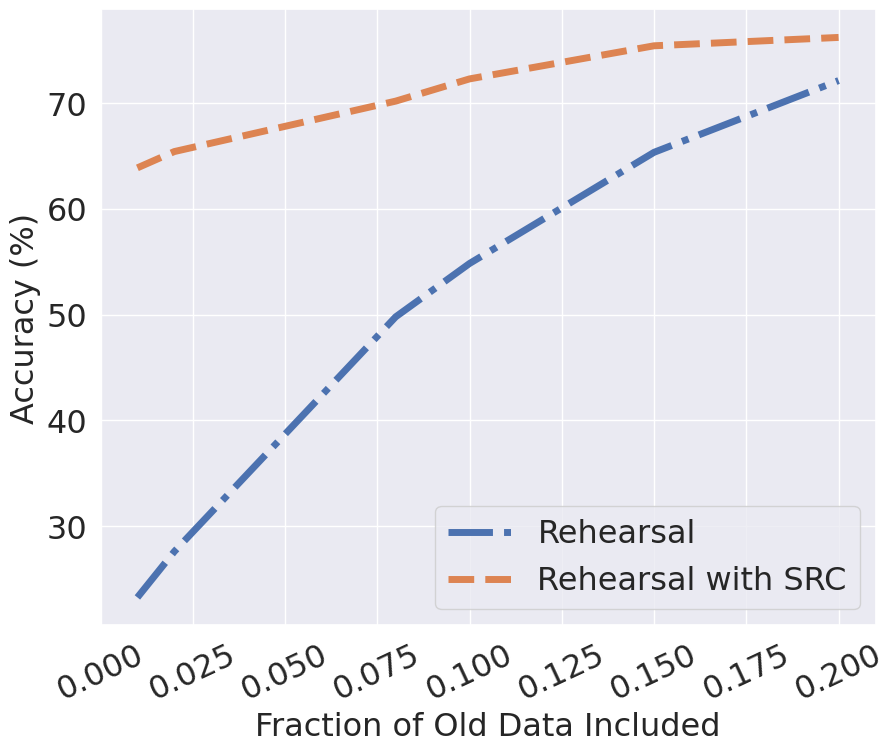}
    \caption{Plot the accuracies of MRNN-EP and the rehearsal method, with and without SRC, as a function of the percentage of old data (rehearsal data) used. The red line represents MRNN-EP combined with SRC and rehearsal, which consistently outperforms the blue line (MRNN-EP with rehearsal but without SRC). This demonstrates the significant advantage of incorporating SRC in enhancing performance, particularly at the final application of SRC and during the last task of the rehearsal method.}
    \label{fig:reh_per_src}
\end{figure}

Figure \ref{fig:reh_per_src} illustrates the results of this analysis for a single task order. When only a small fraction of old data is incorporated, the performance gap between the model with SRC and without SRC is substantial. This demonstrates that SRC significantly enhances the model's ability to retain previously learned tasks, even when minimal old data is available for rehearsal. As amount of old data increased, the gap was narrowed and the models with and without SRC performed similarly when more than ~20\% of old data was used for rehearsal. Importantly, at least 15\% of previously learned data was required to match SRC's performance gain without any old data usage.

Together, these findings suggest that SRC is complimentary to rehearsal and it reduces the dependence on large quantities of old data and training time for maintaining high performance, aligning with observations from previous research \cite{Tadros2022}. By combining equilibrium propagation with SRC, our model exhibits robust continuous learning capabilities, achieving superior performance with a more biologically plausible and data-efficient approach.

\begin{figure}[ht]
\centering
\begin{subfigure}[b]{0.5\textwidth}
   \includegraphics[width=1\linewidth, height=6cm]{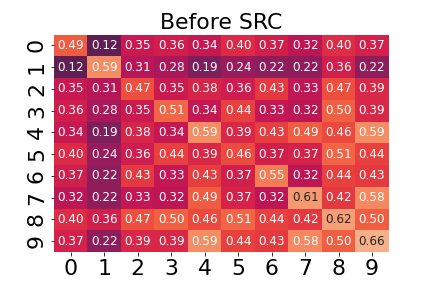}
    \caption{MRNN-EP before SRC}
   \label{fig:before_src} 
\end{subfigure}

\begin{subfigure}[b]{0.5\textwidth}
   \includegraphics[width=1\linewidth, height=6cm]{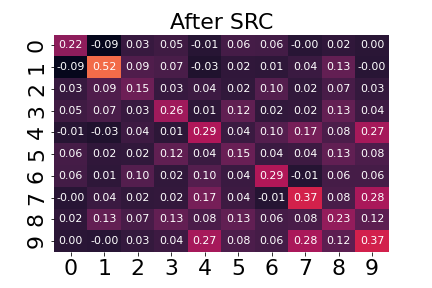}
   \caption{MRNN-EP after SRC}
   \label{fig:after_src}
\end{subfigure}

\caption{Correlation matrix of activations in the hidden layer for the MRNN-EP on the MNIST dataset. After applying SRC, the correlation matrix shows significant changes, with the representations of activations for the classes becoming more independent.}
\end{figure}

\begin{figure*}[ht]
  \begin{subfigure}{0.325\textwidth}
    \includegraphics[width=\linewidth, height=5cm]{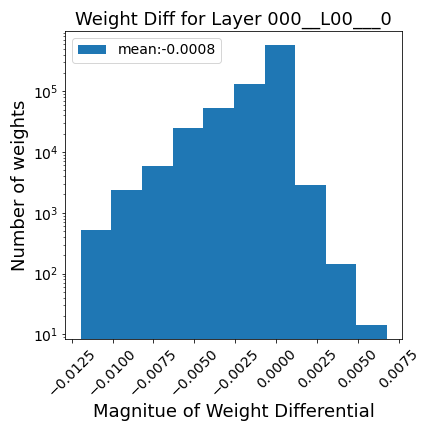}
    \caption{First weights for MRNN-EP after SRC}
    \label{fig:ep_hist_w0}
  \end{subfigure}%
  \hfill
  \begin{subfigure}{0.34\textwidth}
    \includegraphics[width=\linewidth, height=5cm]{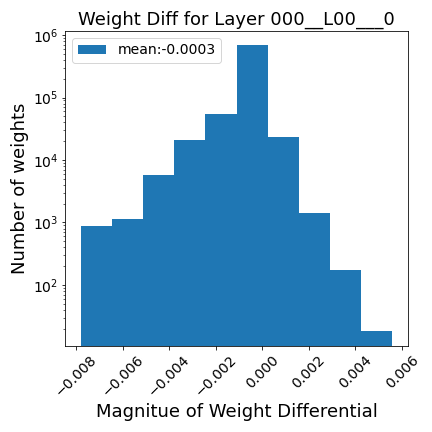}
    \caption{First weights for MRNN-BPTT after SRC}
    \label{fig:bptt_hist_w0}
  \end{subfigure}%
  \hfill
  \begin{subfigure}{0.32\textwidth}
    \includegraphics[width=\linewidth, height=5cm]{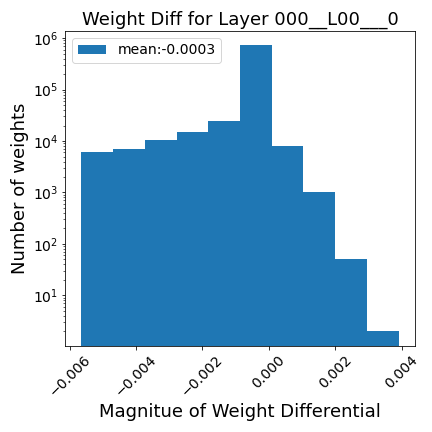}
    \caption{First weights for RNN-BPTT after SRC}
    \label{fig:std_hist_w0}
  \end{subfigure} \\
  
  \begin{subfigure}{0.318\textwidth}
    \includegraphics[width=\linewidth, height=5cm]{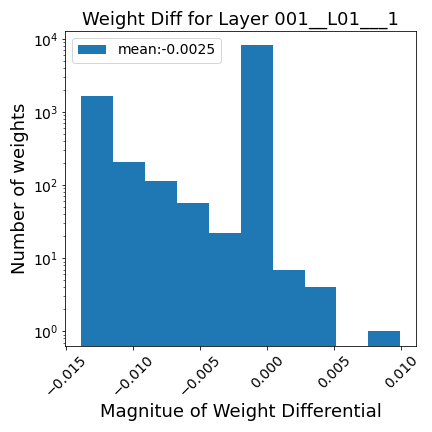}
    \caption{Second weights for MRNN-EP after SRC}
    \label{fig:ep_hist_w1}
  \end{subfigure}%
  \hfill
  \begin{subfigure}{0.338\textwidth}
    \includegraphics[width=\linewidth, height=5cm]{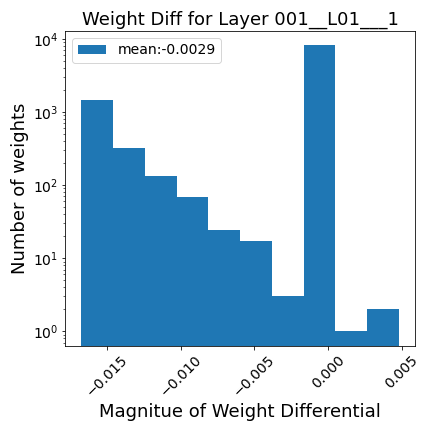}
    \caption{Second weights for MRNN-BPTT after SRC}
    \label{fig:bptt_hist_w1}
  \end{subfigure}%
  \hfill
  \begin{subfigure}{0.33\textwidth}
    \includegraphics[width=\linewidth, height=5cm]{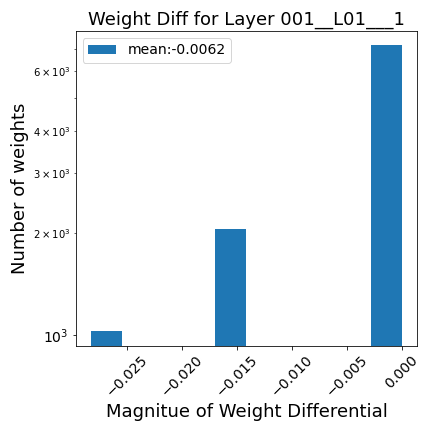}
    \caption{Second weights for RNN-BPTT after SRC}
    \label{fig:std_hist_w1}
  \end{subfigure} \\
  \caption{Histogram of weight differences measured immediately before and after the final SRC phase, which is applied after the last task in the task sequence has been learned (MNIST dataset). ‘First weights’ refer to connections between the input and hidden layers; ‘Second weights’ refer to connections between the hidden and output layers.
Legends indicate the mean values of each histogram. }
  \label{fig:hist_w_mnist}
\end{figure*}

\subsubsection{Correlation matrix of activations in the hidden layer}
To investigate why SRC improves continual learning performance for RNNs trained with EP, we analyzed the correlation matrices of hidden layer activations immediately before and after the final SRC phase, which is applied after the network has finished learning the last task in the sequence. These matrices provide insights into how representations in the hidden layer are shared or separated across dataset classes. Ideally, robust feature learning produces representations that remain largely independent across classes, minimizing task interference and supporting effective retention and generalization.


\Cref{fig:before_src,fig:after_src} depict the correlation matrices of activations in the hidden layer for the MRNN-EP model on the MNIST dataset, prior to and following the last SRC application. Notably, the representations for each class in the hidden layer become substantially more independent in \ref{fig:after_src}, indicating that SRC helps disentangle class-specific features and reduce task interference.

These observations 
suggest that SRC helps allocating distinct neurons to different tasks that would promote independent class representations. Our results suggest that SRC enhances task separation, thereby facilitating robust and efficient learning in continual learning scenarios.

\subsubsection{Analysis of Weight Differences}
We further examined the weight differences across various models prior to and following the last SRC application on the MNIST dataset. Figure \ref{fig:hist_w_mnist} highlights the similarities and differences. In all models the main trend after SRC was shift of majority of weights in negative direction. Notably, in MRNN-EP model, the weights between the input and hidden layers became more negative
following SRC compared to the other models ( \Cref{fig:ep_hist_w0,fig:bptt_hist_w0,fig:std_hist_w0}). This weight dynamics suggests that SRC effectively increases competition between tasks  
by increasing cross-task inhibition, while enhancing only a small fraction of task specific weights.

For the weights between the hidden and output layers, both MRNN-EP and MRNN-BTT revealed similar shift of majority of weights in the negative direction 
(\Cref{fig:ep_hist_w1,fig:bptt_hist_w1,fig:std_hist_w1}). In RNN-BPTT model, decrease of synaptic weights was even more significant  but the changes were grouped in two distinct clusters (\Cref{fig:std_hist_w1}). These weights directly impact the classification process, and excessive and not selective pruning in this layer could potentially remove too many important connections, thereby affecting performance.


\begin{figure*}[ht]
  \begin{subfigure}{0.5\textwidth}
    \includegraphics[width=\linewidth, height=6cm]{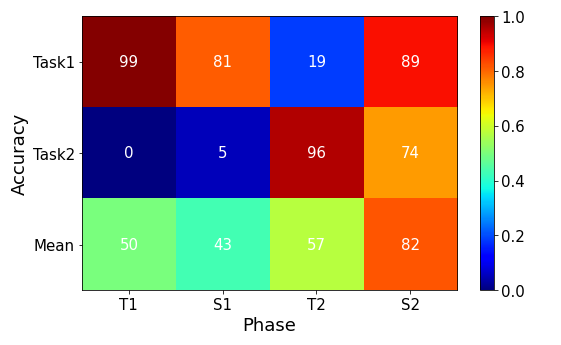}
    \caption{Performance on the simple task.}
    \label{fig:simple_task_perfm}
  \end{subfigure}%
  \hfill
  \begin{subfigure}{0.45\textwidth}
    \includegraphics[width=\linewidth, height=5.5cm]{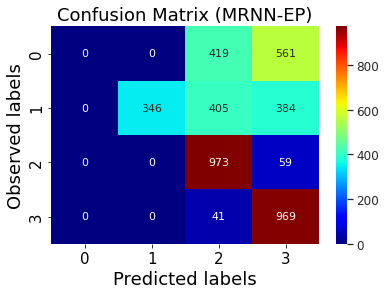}
    \caption{Confusion matrix for Task~1 and Task~2 after Task~2 training (T2). Correct predictions are mostly concentrated on class~1, consistent with the higher synaptic similarity for that class.}
    \label{fig:cm_t2_simple}
  \end{subfigure}%
 \\
  \begin{subfigure}{0.45\textwidth}
    \includegraphics[width=\linewidth, height=4.5cm]{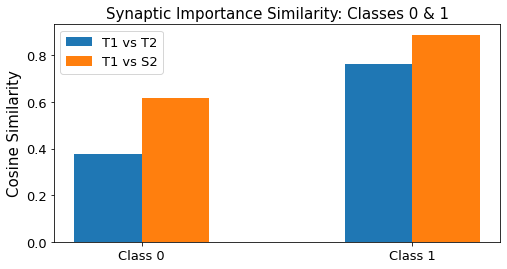}
    \caption{Cosine similarities of synaptic importance patterns for classes 0 and 1: between T1 and T2 (forgetting) and T1 and S2 (recovery)}
    \label{fig:si_cosine_sim_src_01}
  \end{subfigure}%
  \hfill
    \begin{subfigure}{0.45\textwidth}
    \includegraphics[width=\linewidth, height=4.5cm]{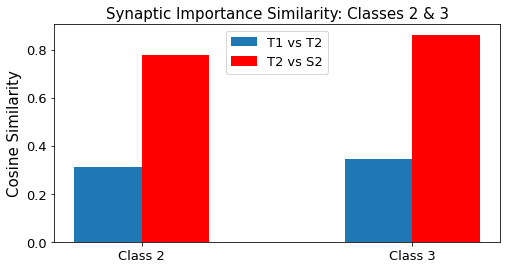}
    \caption{Cosine similarities of synaptic importance patterns for classes 2 and 3: comparing T1 and T2 (emergence of new task representations) and T2 and S2 (preservation of new learning).}
    \label{fig:si_cosine_sim_src_23}
  \end{subfigure}%
\\
  \caption{Synaptic Importance analysis}
  \label{fig:synaptic_importance_analysis}
\end{figure*}


\subsubsection{Analysis of Synaptic Importance on a Simple Task}
To investigate how SRC preserves task-specific knowledge, we conducted a simple analysis using the MNIST dataset, restricted to two tasks: Task~1 (classes 0 and 1) and Task~2 (classes 2 and 3). We trained our MRNN-EP model sequentially on these tasks, applying SRC after each task to examine its effect on synaptic stability. Before presenting the synaptic analysis, we show the performance on this simple task in Figure~\ref{fig:simple_task_perfm}.

After training, we froze the model’s weights and computed the synaptic importance for each class by presenting all corresponding images and measuring the average contribution of each neuron–weight connection:
\begin{equation}
    u_i[c] = \frac{1}{n} \sum_n s_i W_i,
    \label{eq:synaptic_importance}
\end{equation}
where $u_i[c]$ denotes the importance of connection $i$ for class $c$, $s_i$ is the mean activation of neuron $i$ across all samples of class $c$, $W_i$ is the associated weight, and $n$ is the total number of samples for that class. Intuitively, higher $u_i[c]$ values indicate connections that contribute more strongly to representing that specific class.

To assess how SRC influences synaptic patterns, we compared the cosine similarity of these patterns across phases:
\begin{enumerate}
    \item \textbf{T1 vs T2}: For Task~1 classes (classes~0 and~1), this shows how much the original Task~1 synaptic configuration changes after learning Task~2. A lower similarity indicates greater forgetting. For Task~2 classes (classes~2 and~3), this shows how distinct the new task representations are from the prior Task~1 state — lower similarity here reflects the emergence of new task-specific representations.
    \item \textbf{T1 vs S2}: For Task~1 classes, this shows how much of the original Task~1 configuration is recovered after applying SRC following Task~2. A higher similarity suggests that SRC helps restore older task-specific knowledge.
    \item \textbf{T2 vs S2}: For Task~2 classes, this shows how much the new Task~2 configuration is retained after SRC. A high similarity here indicates that SRC preserves new task knowledge while restoring older knowledge.
\end{enumerate}

For this analysis, we focused on the hidden-to-output weights, which directly map learned representations to output classes and thus best reflect class-specific information that is vulnerable to interference.

As shown in Fig.~\ref{fig:si_cosine_sim_src_01}, for classes~0 and~1 the cosine similarity between T1 and S2 is consistently higher than that between T1 and T2, indicating that SRC helps restore older synaptic configurations that would otherwise be overwritten by new learning. Specifically, for class~0 the similarity increased from 0.38 (T1 vs T2) to 0.62 (T1 vs S2), while for class~1 it increased from 0.76 (T1 vs T2) to 0.89 (T1 vs S2).

A similar trend appears for classes~2 and~3, shown in Fig.~\ref{fig:si_cosine_sim_src_23}. Here, the similarity between T2 and S2 is markedly higher than the similarity between T1 and T2. For example, for class~2, the similarity jumps from 0.31 (T1 vs T2) to 0.78 (T2 vs S2), and for class~3, from 0.35 (T1 vs T2) to 0.86 (T2 vs S2). These results further confirm that SRC selectively recovers older representations disrupted by new tasks.

Interestingly, for class~1, the similarity between T1 and T2 was already relatively high even before applying SRC (0.76). This indicates that training on the new task did not completely overwrite the original class~1 pattern, leaving behind a residual trace in the synaptic configuration. This residual similarity explains why Task~1 performance did not drop entirely to chance after Task~2 training but remained at approximately 19\% (please check the performance in Figure~\ref{fig:simple_task_perfm}). To illustrate this effect, we examined the confusion matrix for Task~1 after Task~2 training (T2 phase), shown in Fig.~\ref{fig:cm_t2_simple}. It confirms that this residual accuracy is almost entirely due to correct predictions for class~1, while class~0 was completely forgotten. This highlights SRC’s benefit: it restores broader task-specific patterns beyond what remains through incomplete overwriting alone.

\section{Discussion}

In this work, we introduced and applied an unsupervised, sleep-inspired replay algorithm—Sleep Replay Consolidation (SRC)—to multi-layer recurrent neural networks trained with equilibrium propagation (MRNN-EP). Our results show that SRC significantly enhances MRNN-EP performance in class-incremental learning settings. While SRC yielded comparable improvements for MRNN-BPTT on the MNIST dataset, MRNN-EP consistently outperformed MRNN-BPTT and RNN-BPTT on more challenging datasets, including Fashion MNIST, Kuzushiji-MNIST, CIFAR-10, and ImageNet, demonstrating the potential of SRC in advancing biologically plausible learning. Moreover, combining SRC with rehearsal methods produced complementary effects, further boosting performance and effectively mitigating catastrophic forgetting.


Our specific contributions are as follows: 
\begin{enumerate} 
\item Sleep Replay Consolidation: We propose a novel, biologically inspired optimization algorithm called Sleep Replay Consolidation (SRC) for RNNs trained using EP. SRC emulates sleep replay in the biological brain to mitigate catastrophic forgetting and improve retention in continual learning tasks.

\item Comparative Evaluation: We provide a comprehensive evaluation of SRC's benefits for RNNs trained with EP versus those trained with Backpropagation Through Time (BPTT). Our findings indicate that while SRC provides similar performance improvements to both models on simpler tasks, as a local learning rule, SRC may align better with the local nature of EP plasticity, 
providing beneficial for more complex tasks.


\item Integration with Rehearsal: By combining SRC with rehearsal methods, we demonstrate further improvements in continual learning, showcasing enhanced long-term retention and data efficiency. 
\end{enumerate}


The critical role that sleep plays in learning and memory is supported by a vast, interdisciplinary literature spanning both psychology and neuroscience \cite{Paller2004,Walker2004,Oudiette2013,Rasch2013,Stickgold2013b}. While the “classic” picture suggests that, in vertebrates, Rapid-Eye-Movement (REM) sleep supports the consolidation of non-declarative or procedural memories, and non-REM sleep supports the consolidation of declarative memories \cite{Mednick2011,Rasch2013,Stickgold2013b,Ramanathan2015}, it becomes increasingly evident that such separation is oversimplified, and both types of sleep may be needed for learning complex and potentially conflicting tasks. Some theories suggest that NREM sleep has a role in strengthening of recent memory traces whereas REM sleep is important for synaptic rebalancing and pruning to eliminate unnecessary traces \cite{O2014,Li2017}. Despite the difference in the cellular and network dynamics during these two stages of sleep \cite{Rasch2013,Stickgold2013b}, both are thought to contribute to memory consolidation through repeated reactivation, or replay, of specific memory traces acquired during learning \cite{Hennevin1995,Paller2004,Mednick2011,Oudiette2013,Rasch2013,Lewis2018,Wei2018}.

The model of sleep provided by SRC resembles REM sleep, as it lacks the clear large-scale synchronization dynamics typical of NREM sleep, such as sleep slow oscillation \cite{Steriade1993}. Our study predicts that the main effect of REM sleep is an increase in inhibitory effects, which is consistent with the known literature on the synaptic effects of REM \cite{O2014,Li2017}.
Our results highlight that these biologically inspired strategies are pivotal in enhancing the capabilities of RNNs trained with EP, suggesting that they help make RNNs more biologically plausible. 


While the performance of MRNN-EP with SRC approaches that of models trained in parallel, a notable performance decline was observed when raw inputs were used instead of features extracted from pretrained networks. That is, our implementation still falls short of matching the human brain's remarkable ability to retain previous knowledge. This indicates that some critical mechanisms inherent to biological systems are still missing in our model, underscoring the need for further research to bridge the gap between artificial and biological learning systems.

Applications of SRC are not limited to continual learning. Recent studies \cite{Delanois2023} suggest that SRC can enhance generalization in convolutional neural networks, presenting a compelling avenue for further exploration. Another study revealed that SRC can improve model performance when training data are limited or unbalanced \cite{Bazhenov2024}. An unexplored but exciting direction involves applying SRC to neural networks trained with EP in reinforcement learning frameworks. This extension could provide biologically inspired solutions for dynamic, task-driven environments, opening new possibilities for continual learning in real-world applications.


Since SRC is implemented within the spiking neural network (SNN) domain, it may be beneficial to combine SRC with SNNs trained using EP. Indeed, recent research \cite{Connor2019,Martin2021,Taylor2022,Lin2024} has demonstrated the feasibility of EP for training SNNs. Leveraging SRC’s benefits alongside the energy efficiency and event-driven computation of SNNs could pave the way for significant advancements in biologically plausible continuous learning systems.

\section*{Acknowledgments}
This study is supported by NSF (grants 2223839 and 2323241) and NIH (grants R01MH125557 and RF1NS132913).

\bibliographystyle{named}
\bibliography{ijcai24}

\end{document}